\documentclass[sigconf]{acmart}

\usepackage{todonotes}

\AtBeginDocument{%
  \providecommand\BibTeX{{%
    \normalfont B\kern-0.5em{\scshape i\kern-0.25em b}\kern-0.8em\TeX}}}

\copyrightyear{2021}
\acmYear{2021}
\acmConference[WWW '21 Companion]{Companion Proceedings of the Web Conference 2021}{April 19--23, 2021}{Ljubljana, Slovenia}
\acmBooktitle{Companion Proceedings of the Web Conference 2021 (WWW '21 Companion), April 19--23, 2021, Ljubljana, Slovenia}
\acmPrice{}
\acmDOI{10.1145/3442442.3452337}
\acmISBN{978-1-4503-8313-4/21/04}

\begin{document}

\title{References in Wikipedia: The Editors’ Perspective}

\author{Lucie-Aim\'{e}e Kaffee}
\orcid{0000-0002-1514-8505}
\affiliation{%
  \institution{University of Southampton}
  \city{}
  \country{}
}
\email{kaffee@soton.ac.uk}

\author{Hady Elsahar}
\affiliation{
  \institution{Naver Labs Europe}
  \city{}
  \country{}
  }
\email{hady.elsahar@naverlabs.com}

%\renewcommand{\shortauthors}{Kaffee et al.}

\begin{abstract}
  References are an essential part of Wikipedia. Each statement in Wikipedia should be referenced. In this paper, we explore the creation and collection of references for new Wikipedia articles from an editor's perspective. We map out the workflow of editors when creating a new article, emphasising on how they select references.
\end{abstract}


\maketitle

\section{Introduction}
% WHY WHAT HOW
Wikipedia is one of the most used sources for information online.
As Wikipedia understands itself as a encyclopedia, it is a secondary source, i.e., it does not contain original research but all statements are referenced to external sources. 
Wikipedia is written and maintained by a community of editors, who ensure that the quality of the articles can be maintained and add references to articles.
Currently, we do not know what is the workflow of an editor when they create an article, with a focus on the references they are using and when they are introducing references to their workflow.
Therefore we conducted two studies to gain an insight into how editors create articles with respect to referencing.

The study is twofold: First, we conducted a survey to which tools editors are using to create articles and find references.
Then, we conducted a series of interviews with Wikipedia editors across five very differently covered language versions to understand the workflow of article creating and referencing in more depth.

We find that references are an essential part of the editors' workflows to an extend that they structure their work around it, such as the structuring of the article they are writing.

\section{Related Work}
Wikipedia's main principle is that anyone can edit and contribute to the encyclopedia. 
Editors collaborate and coordinate through talk pages \cite{DBLP:conf/hicss/ViegasWKH07}.
In our work, we study how editors create articles with a focus on how they use references.

References in Wikipedia are an essential part of the encyclopedia to ensure the quality of the information.
Similar to our work, \citet{DBLP:journals/ires/Huvila10} conducts surveys to understand the provenance of information on Wikipedia. They conclude that editors favour online sources to discover information. They distinguish five editor types: Investigators, Surfers, Worldly-wise, Scholars, and Editors. In their work, they also find that editors usually write articles in their area of expertise. Web sources are named as the most common source to discover information, followed by offline resources in the form of books.
We extend their study by focusing on tools editors use to create articles in the survey, and interview editors in semi-structured interviews.
\citet{DBLP:journals/corr/abs-2002-04347} analyse reference usage of editors in the science domain by analysing existing references.

Regarding the reuse of Wikipedia references, \citet{DBLP:conf/www/PiccardiRC020} explore the interaction with references by readers of Wikipedia. They find 93\% of online references are never clicked on in English Wikipedia, i.e., interaction with references on Wikipedia is low.

A large body of work is concerned with the quality and usage of references in Wikipedia.
\citet{DBLP:journals/information/LewoniewskiWA20} propose models to assess popularity and reliability of references in Wikipedia. 
\citet{DBLP:journals/corr/abs-2007-07022} provide a dataset of English Wikipedia references, focusing on scholarly publications.
\citet{DBLP:conf/icist/LewoniewskiWA17} analyse usage of references on Wikipedia across multiple language versions.

\section{Methods}
We employed a mixed-method approach, in which we created a survey to collect structured information about how people are adding references to their Wikipedia when writing an article and conducted a series of semi-structured interviews to get an insight view on the editors’ editing. The survey was a Google survey. For the interviews, we recruited participants in Wikimedia events (such as the Wikimedia Hackathon 2019 in Prague) and via Wikimedia mailing lists. 
The survey and interviews were conducted in English.

\section{Survey}

The surveys aimed to collect information about what tools Wikipedia editors already use in their work of article creation and finding references.
All answers could be input into free text fields.
We asked two demographics questions, which Wikipedia they edit (including the edit count), and since when they edit Wikipedia.
Then, we asked them the two questions that are aligned with our overall research question: which tools they use for article creation and which tools they use for finding references. 
We coded the answers from the free-text fields for the last two questions and grouped them by type of task solved for each set of tools.

\paragraph{\textbf{Demographics}}
We collected 19 responses from editors across different Wikipedia language versions over the course of 2019. Editors edit between one and 28 Wikipedia language versions. Editors have between 2 and 55,000 edits on these Wikipedias.

\begin{figure}
    \centering
    \includegraphics[width=\columnwidth]{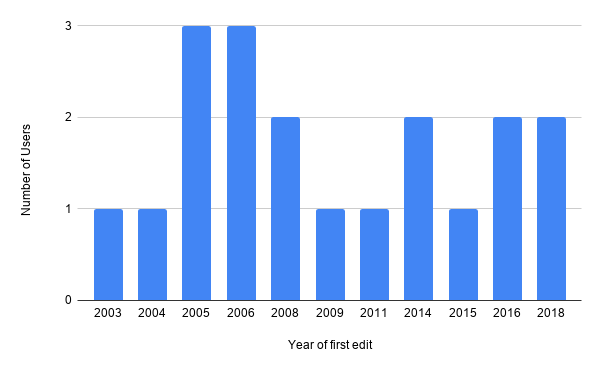}
    \caption{Distribution of editors’ “seniority”, i.e., the years they first started editing Wikipedia.}
    \label{fig:editors_survey}
\end{figure}

The majority of the editors have been editing over two years, see Figure~\ref{fig:editors_survey}. The median starting year is 2008. Given English Wikipedia started in 2001, we can say the editors who filled in the survey are rather experienced. 
\paragraph{\textbf{Tools for Article Creation}} \textit{32\% of editor do not use tools for article creation. The tools used are for preparation and formatting text, content structuring, and for advice on how to write articles.}

We asked the editors which tools they use to create new articles. This gives an insight into the usage of tools but also integrates with the follow-up question about what tools they use to find references for their articles.

Six editors say they do not use any specific tools to create new articles. Among the tools editors use, are tools that help them in text formatting and preparation of the article writing, such as Libre Office, Google docs (as Wikipedia does not let one save a draft without publishing), text editor and clipboard manager, and Visual editor, a Wikipedia-internal rich text editor. To find advice on how to create an article, editors would ask peers and Wiki guidance, i.e., pages on Wikipedia for more general Wikipedia guidelines. Two editors state they use the ContentTranslation tool, while another editor says they look at existing articles about similar topics and adapt from those. Further, in the realm of content structuring, editors use excel and not further specified self-made tools, as well as online resources such as Google search and online databases. 

This wide range of tools indicates how experienced editors have built their own editing routines and found the tools to support them. However, there are still some valuable lessons from their article creation to routines: There is a need for a good editing interface, which has been partially addressed with the Visual editor. Further, guidance on how to write an article based on existing articles is still wanted even by more experienced editors. 

\paragraph{\textbf{Tools for Finding References}} \textit{11\% of editors do not use any tools for finding references. 5\% use only offline resources, 11\% use online and offline resources. Online sources range from general search engines to academic resources and English sources.}

In the next part of the survey, we ask editors what tools they use for finding references for an article they are writing. Two editors say they do not use any tools, which seems to be because the word tool has a very specific Wikipedia-internal context. In the other editors, there are editors who use online and offline resources. One editor states they only use offline literature (i.e., from the library), the other two editors using offline references such as books and newspapers use them together with online resources, such as Google search. One editor states they only use Google, while two editors use only academic online resources, such as academic databases and Google scholar. Other editors state they look specifically for online databases on the topic they write about. One editor states they mostly use existing references from English Wikipedia in their Wikipedia. However, they do not state they use the ContentTranslation tool in the previous question but rather have two browser windows for the article they write. 

\section{Semi-Structured Interviews}
We created a set of semi-structured interview questions to explore referencing in Wikipedia in-depth. Our emphasis was to understand ways to gather citable sources when writing a Wikipedia article. Two of the interviews were conducted online, three at in-person events (the Wikimedia Hackathon 2019\footnote{\url{https://www.mediawiki.org/wiki/Wikimedia_Hackathon_2019}}). 
The research was approved by the Ethics Committee of the University of Southampton under ERGO Number 48849.

\begin{figure}
    \centering
    \includegraphics[width=\columnwidth]{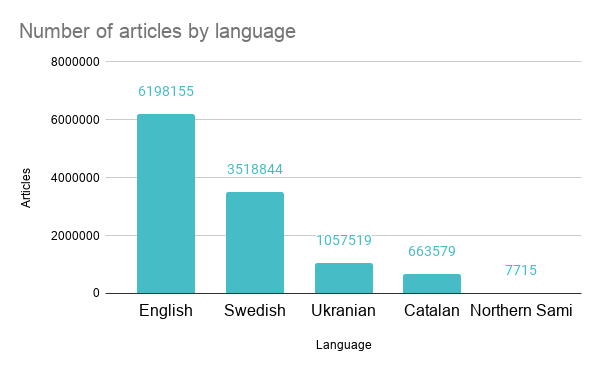}
    \caption{Number of articles in the Wikipedia languages of the interview participants.}
    \label{fig:interview_language}
\end{figure}

\begin{figure}
    \centering
    \includegraphics[width=\columnwidth]{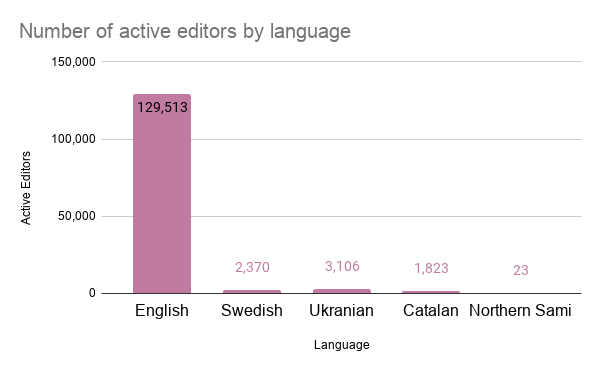}
    \caption{Number of active editors in the Wikipedia languages of the participants.}
    \label{fig:interview_active_editors}
\end{figure}

\paragraph{\textbf{Demographics}}
We selected the participants to cover a broad range of Wikipedia-size. From English Wikipedia with the most articles as of now, to Swedish with a large number of bot-generated articles, and Ukranian and Catalan, and finally Northern Sami with a small number of articles and a relatively small community, see Figure~\ref{fig:interview_language}. In Figure \ref{fig:interview_active_editors}, we can see the number of articles and editors for those Wikipedia languages as of November 2020. All editors are experienced editors and have been contributing to Wikipedia for over five years.

\paragraph{\textbf{Selection of topic}}\textit{Topic are selected based on personal interest or community need.}

First, we explore how editors pick topics to write an article on. All editors select topics they are interested in creating, i.e., from their own field of interest, which aligns with the findings of \citet{DBLP:journals/ires/Huvila10}. They all tell stories about experts in their Wikipedia that have created many articles in one field of their Wikipedia. Especially in very small Wikipedias such as Nothern Sami, this leads to a high coverage of one topic (such as birds) while not yet having more basic concepts (such as \textit{hand}). For some editors, it is important that their language does not have a lot of books on a topic, and Wikipedia can fill this gap in educational content in their language. To identify articles that are wanted by the communities, editors have different strategies. Larger Wikipedia communities have campaigns for topics, including red links of missing articles in a topic area. Additionally or for editors of smaller Wikipedia communities, they might look at social media, ask teachers what their students miss in content in their language, or look at page views.

\begin{figure}
    \centering
    \includegraphics[width=\columnwidth]{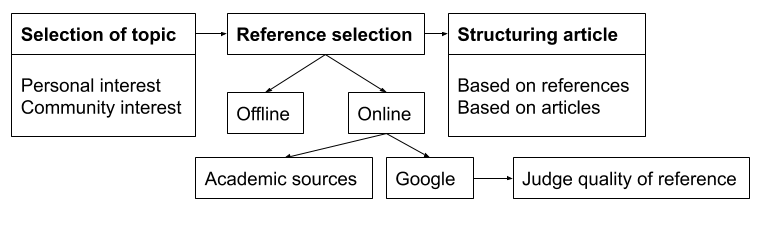}
    \caption{Most common workflow among the editors interviewed}
    \label{fig:interview_workflow}
\end{figure}

\paragraph{\textbf{Workflows}}\textit{There are two main workflows: based on references or on article sections, a summary of the most common workflow can be found in Figure~\ref{fig:interview_workflow}.}

We identified two main workflows after selecting a topic to write an article about. The most common approach among the interviewed participants was to first look for references. Then, they might sort the references to create the sections, or start writing the article in their userspace. Often one of the considerations from the beginning is on how to convince other editors, that their topic is notable, especially in larger Wikipedia communities such as English Wikipedia. 
The other approach to write an article is to first order the article in section and then look for references for each section, one after another, starting with the definition on the top part of the article.
If a user speaks multiple languages, they might create the article simultaneously in all the languages they speak. All non-English Wikipedia editors report that especially content specific for their cultural context is not covered in other languages, therefore not being able to translate it. Bots in Swedish Wikipedia lead to a high number of articles while there is still a lack in content. For English Wikipedia in comparison, the editor is mass-creating articles based on databases which then can be edited by humans, leading to a high number of articles created and based on experience, enriching those with more content. 

\paragraph{\textbf{Section Structure}}\textit{The structure of the article is created based on experience, references, or guidelines and existing articles.}

To create the structure of the article, editors report that it depends on the topic they are writing about. Most editors say they use a standard of Wikipedia. Interestingly, this standard of section structure is not defined but ``a template in the head'' of the editor, i.e., they decide on the section structure based on experience. One editor says, is is a mix of experience and looking at the references at hand, retrieved in their previous step. Less experienced editors or when working on a new topic, they will look at featured articles or guidelines. While most editors feel like the structure of the articles in their topic of expertise is fairly standard across different language Wikipedias, one editor says the structure can differ based on cultural differences, e.g., English Wikipedia would have a section on charity activities on the end of contemporary biographies, but Swedish Wikipedia does not as it is not considered an essential part in a biography in an encyclopedia. One editor reports they might change the section structure while writing as there is no standard section structure in their language Wikipedia due to the low number of articles overall. 

\paragraph{\textbf{Online References}}\textit{Online references are chosen based on accessibility, low resourced languages have a lack of offline resources in their language. Quality of reference can be inferred by previous usage of the publisher on Wikipedia.}

While many editors use offline resources for their articles, many editors report also the use of online references in their work. One editor mainly uses offline references when writing on articles they have less expertise and therefore less access to offline references. 
Online references are accessed through search engines, by looking for existing resources written about the topic in the Wikipedia’s language. If the editor is knowledgeable in the topic, they may resort directly to databases or academic resources specifically for the topic they are writing about. A big barrier here is the inaccessibility of academic sources from outside academia, making exceptions such as jstore’s three free articles per month a highlight for the editors even when being part of a university.

Catalan and Northern Sami have a limited number of offline references overall, leaving them to resort mostly to online references. 
For the languages with limited resources available, overall there is a problem of finding references in their language, which might lead them to looking for references in other languages to enrich the quality of the article. One editor takes this an advantage, suggesting a more diverse set of references, also in terms of languages, enables them to discuss a topic from multiple perspectives to create a more objective article.

When there is a low number of resources, a good indicator for the quality of a reference can be whether this reference has been used on Wikipedia before. But also for larger languages, if the editor is not familiar with the publisher, they might investigate if this reference has been used before on Wikipedia.

\section{Discussion and Conclusion}
In this paper, we explore Wikipedia editors' workflow with a focus on referencing when writing new articles.
We describe the editors' workflow in a two fold manner: First, we create a survey, and then we interview a set of editors across different language versions of Wikipedia.
First, we establish that experienced editors use tools to support them in their work. However, those are not unified, i.e., different editors use different tools for the same or different tasks. 

Then, we dive into the editors' workflow. Editors select a topic to edit based on personal or the communities' interest. 
The availability of references influences the way they write articles.
Given most editors use online resources as references, we believe the process of creating references can be supported by filtering the references that can be retrieved online. 

The structure of articles of Wikipedia is an important, often underestimated, factor for the readers of Wikipedia and influence their interactions with the articles \cite{DBLP:journals/nrhm/LamprechtLHS17}.
Currently, editors find ways to structure articles based on experience, the references available to them or guidelines. Future work can support editors in creating better article structure by making it easier to find a fitting structure by making it easier to access common structures of the articles in their language Wikipedia while writing an article.

The quality of references is an important aspect of referencing in Wikipedia. We find one criteria for editors, beside their experience, is whether a reference is used before on their or English Wikipedia.
Important to emphasise here is also the lack of references for lower-resourced languages. \citet{DBLP:journals/jasis/TeplitskiyLD17} find that open access journals are more likely to be cited on Wikipedia, which emphasises the importance of accessibility of academic research.
Wikipedia is the most read online encyclopedia, ensuring access to knowledge on a broad range of topics in a variety of languages. Therefore, it is important that editors are able to access original research for the quality-assurance of their articles.

This work aims at giving an insight into the editors' workflows, with a focus on references, to give a starting point for future work on supporting editors in creating articles and referencing.
\begin{acks}
This work was supported by the Wikimedia Foundation Project Grant under the title \textit{Scribe: Supporting Under-resourced Wikipedia Editors in Creating New Articles}, \url{https://meta.wikimedia.org/wiki/Scribe}. The authors thank the Wikimedia community members, who participated in the survey and interviews.
\end{acks}

\bibliographystyle{ACM-Reference-Format}
\bibliography{acmart}


\begin{thebibliography}{9}

%%% the Web address.

\ifx \showCODEN    \undefined \def \showCODEN     #1{\unskip}     \fi
\ifx \showDOI      \undefined \def \showDOI       #1{#1}\fi
\ifx \showISBNx    \undefined \def \showISBNx     #1{\unskip}     \fi
\ifx \showISBNxiii \undefined \def \showISBNxiii  #1{\unskip}     \fi
\ifx \showISSN     \undefined \def \showISSN      #1{\unskip}     \fi
\ifx \showLCCN     \undefined \def \showLCCN      #1{\unskip}     \fi
\ifx \shownote     \undefined \def \shownote      #1{#1}          \fi
\ifx \showarticletitle \undefined \def \showarticletitle #1{#1}   \fi
\ifx \showURL      \undefined \def \showURL       {\relax}        \fi
% invisible to TeX
\providecommand\bibfield[2]{#2}
\providecommand\bibinfo[2]{#2}
\providecommand\natexlab[1]{#1}
\providecommand\showeprint[2][]{arXiv:#2}

\bibitem[\protect\citeauthoryear{Arroyo{-}Machado, Torres{-}Salinas,
  Herrera{-}Viedma, and Romero{-}Fr{\'{\i}}as}{Arroyo{-}Machado
  et~al\mbox{.}}{2020}]%
        {DBLP:journals/corr/abs-2002-04347}
\bibfield{author}{\bibinfo{person}{Wenceslao Arroyo{-}Machado},
  \bibinfo{person}{Daniel Torres{-}Salinas}, \bibinfo{person}{Enrique
  Herrera{-}Viedma}, {and} \bibinfo{person}{Esteban Romero{-}Fr{\'{\i}}as}.}
  \bibinfo{year}{2020}\natexlab{}.
\newblock \showarticletitle{Science through Wikipedia: {A} novel representation
  of open knowledge through co-citation networks}.
\newblock \bibinfo{journal}{\emph{CoRR}}  \bibinfo{volume}{abs/2002.04347}
  (\bibinfo{year}{2020}).
\newblock
\showeprint[arxiv]{2002.04347}
\urldef\tempurl%
\url{https://arxiv.org/abs/2002.04347}
\showURL{%
\tempurl}


\bibitem[\protect\citeauthoryear{Huvila}{Huvila}{2010}]%
        {DBLP:journals/ires/Huvila10}
\bibfield{author}{\bibinfo{person}{Isto Huvila}.}
  \bibinfo{year}{2010}\natexlab{}.
\newblock \showarticletitle{Where does the information come from? Information
  source use patterns in Wikipedia}.
\newblock \bibinfo{journal}{\emph{Inf. Res.}} \bibinfo{volume}{15},
  \bibinfo{number}{3} (\bibinfo{year}{2010}).
\newblock
\urldef\tempurl%
\url{http://www.informationr.net/ir/15-3/paper433.html}
\showURL{%
\tempurl}


\bibitem[\protect\citeauthoryear{Lamprecht, Lerman, Helic, and
  Strohmaier}{Lamprecht et~al\mbox{.}}{2017}]%
        {DBLP:journals/nrhm/LamprechtLHS17}
\bibfield{author}{\bibinfo{person}{Daniel Lamprecht}, \bibinfo{person}{Kristina
  Lerman}, \bibinfo{person}{Denis Helic}, {and} \bibinfo{person}{Markus
  Strohmaier}.} \bibinfo{year}{2017}\natexlab{}.
\newblock \showarticletitle{How the structure of Wikipedia articles influences
  user navigation}.
\newblock \bibinfo{journal}{\emph{New Rev. Hypermedia Multim.}}
  \bibinfo{volume}{23}, \bibinfo{number}{1} (\bibinfo{year}{2017}),
  \bibinfo{pages}{29--50}.
\newblock
\urldef\tempurl%
\url{https://doi.org/10.1080/13614568.2016.1179798}
\showDOI{\tempurl}


\bibitem[\protect\citeauthoryear{Lewoniewski, Wecel, and
  Abramowicz}{Lewoniewski et~al\mbox{.}}{2017}]%
        {DBLP:conf/icist/LewoniewskiWA17}
\bibfield{author}{\bibinfo{person}{Wlodzimierz Lewoniewski},
  \bibinfo{person}{Krzysztof Wecel}, {and} \bibinfo{person}{Witold
  Abramowicz}.} \bibinfo{year}{2017}\natexlab{}.
\newblock \showarticletitle{Analysis of References Across Wikipedia Languages}.
  In \bibinfo{booktitle}{\emph{Information and Software Technologies - 23rd
  International Conference, {ICIST} 2017, Druskininkai, Lithuania, October
  12-14, 2017, Proceedings}} \emph{(\bibinfo{series}{Communications in Computer
  and Information Science}, Vol.~\bibinfo{volume}{756})},
  \bibfield{editor}{\bibinfo{person}{Robertas Damasevicius} {and}
  \bibinfo{person}{Vilma Mikasyte}} (Eds.). \bibinfo{publisher}{Springer},
  \bibinfo{pages}{561--573}.
\newblock
\urldef\tempurl%
\url{https://doi.org/10.1007/978-3-319-67642-5\_47}
\showDOI{\tempurl}


\bibitem[\protect\citeauthoryear{Lewoniewski, Wecel, and
  Abramowicz}{Lewoniewski et~al\mbox{.}}{2020}]%
        {DBLP:journals/information/LewoniewskiWA20}
\bibfield{author}{\bibinfo{person}{Wlodzimierz Lewoniewski},
  \bibinfo{person}{Krzysztof Wecel}, {and} \bibinfo{person}{Witold
  Abramowicz}.} \bibinfo{year}{2020}\natexlab{}.
\newblock \showarticletitle{Modeling Popularity and Reliability of Sources in
  Multilingual Wikipedia}.
\newblock \bibinfo{journal}{\emph{Inf.}} \bibinfo{volume}{11},
  \bibinfo{number}{5} (\bibinfo{year}{2020}), \bibinfo{pages}{263}.
\newblock
\urldef\tempurl%
\url{https://doi.org/10.3390/info11050263}
\showDOI{\tempurl}


\bibitem[\protect\citeauthoryear{Piccardi, Redi, Colavizza, and West}{Piccardi
  et~al\mbox{.}}{2020}]%
        {DBLP:conf/www/PiccardiRC020}
\bibfield{author}{\bibinfo{person}{Tiziano Piccardi}, \bibinfo{person}{Miriam
  Redi}, \bibinfo{person}{Giovanni Colavizza}, {and} \bibinfo{person}{Robert
  West}.} \bibinfo{year}{2020}\natexlab{}.
\newblock \showarticletitle{Quantifying Engagement with Citations on
  Wikipedia}. In \bibinfo{booktitle}{\emph{{WWW} '20: The Web Conference 2020,
  Taipei, Taiwan, April 20-24, 2020}},
  \bibfield{editor}{\bibinfo{person}{Yennun Huang}, \bibinfo{person}{Irwin
  King}, \bibinfo{person}{Tie{-}Yan Liu}, {and} \bibinfo{person}{Maarten van
  Steen}} (Eds.). \bibinfo{publisher}{{ACM} / {IW3C2}},
  \bibinfo{pages}{2365--2376}.
\newblock
\urldef\tempurl%
\url{https://doi.org/10.1145/3366423.3380300}
\showDOI{\tempurl}


\bibitem[\protect\citeauthoryear{Singh, West, and Colavizza}{Singh
  et~al\mbox{.}}{2020}]%
        {DBLP:journals/corr/abs-2007-07022}
\bibfield{author}{\bibinfo{person}{Harshdeep Singh}, \bibinfo{person}{Robert
  West}, {and} \bibinfo{person}{Giovanni Colavizza}.}
  \bibinfo{year}{2020}\natexlab{}.
\newblock \showarticletitle{Wikipedia Citations: {A} comprehensive dataset of
  citations with identifiers extracted from English Wikipedia}.
\newblock \bibinfo{journal}{\emph{CoRR}}  \bibinfo{volume}{abs/2007.07022}
  (\bibinfo{year}{2020}).
\newblock
\showeprint[arxiv]{2007.07022}
\urldef\tempurl%
\url{https://arxiv.org/abs/2007.07022}
\showURL{%
\tempurl}


\bibitem[\protect\citeauthoryear{Teplitskiy, Lu, and Duede}{Teplitskiy
  et~al\mbox{.}}{2017}]%
        {DBLP:journals/jasis/TeplitskiyLD17}
\bibfield{author}{\bibinfo{person}{Misha Teplitskiy}, \bibinfo{person}{Grace
  Lu}, {and} \bibinfo{person}{Eamon Duede}.} \bibinfo{year}{2017}\natexlab{}.
\newblock \showarticletitle{Amplifying the impact of open access: Wikipedia and
  the diffusion of science}.
\newblock \bibinfo{journal}{\emph{J. Assoc. Inf. Sci. Technol.}}
  \bibinfo{volume}{68}, \bibinfo{number}{9} (\bibinfo{year}{2017}),
  \bibinfo{pages}{2116--2127}.
\newblock
\urldef\tempurl%
\url{https://doi.org/10.1002/asi.23687}
\showDOI{\tempurl}


\bibitem[\protect\citeauthoryear{Vi{\'{e}}gas, Wattenberg, Kriss, and van
  Ham}{Vi{\'{e}}gas et~al\mbox{.}}{2007}]%
        {DBLP:conf/hicss/ViegasWKH07}
\bibfield{author}{\bibinfo{person}{Fernanda~B. Vi{\'{e}}gas},
  \bibinfo{person}{Martin Wattenberg}, \bibinfo{person}{Jesse Kriss}, {and}
  \bibinfo{person}{Frank van Ham}.} \bibinfo{year}{2007}\natexlab{}.
\newblock \showarticletitle{Talk Before You Type: Coordination in Wikipedia}.
  In \bibinfo{booktitle}{\emph{40th Hawaii International International
  Conference on Systems Science {(HICSS-40} 2007), {CD-ROM} / Abstracts
  Proceedings, 3-6 January 2007, Waikoloa, Big Island, HI, {USA}}}.
  \bibinfo{publisher}{{IEEE} Computer Society}, \bibinfo{pages}{78}.
\newblock
\urldef\tempurl%
\url{https://doi.org/10.1109/HICSS.2007.511}
\showDOI{\tempurl}


\end{thebibliography}

\end{document}